\documentclass[11pt,letterpaper]{article}
\usepackage{emnlp2017}
\usepackage{times}
\usepackage{latexsym}
\usepackage{subfigure}
\usepackage{graphicx}
\usepackage{linguex}
\definecolor{light-gray}{gray}{0.7}
\usepackage{lipsum}
\emnlpfinalcopy

\def\emnlppaperid{***}

\newcommand\blfootnote[1]{%
  \begingroup
  \renewcommand\thefootnote{}\footnote{#1}%
  \addtocounter{footnote}{-1}%
  \endgroup
}

\title{Improving coreference resolution\\ with automatically predicted prosodic information}

\author{Ina R\"osiger$^*$, Sabrina Stehwien$^*$, \textbf{Arndt Riester, Ngoc Thang Vu}
\\ Institute for Natural Language Processing \\University of Stuttgart, Germany \\
  {\tt \{roesigia,stehwisa,arndt,thangvu\}@ims.uni-stuttgart.de}}

\date{}

\begin{document}

\maketitle

\begin{abstract}
Adding manually annotated prosodic information, specifically pitch accents and phrasing, to the typical text-based feature set for coreference resolution has previously been shown to have a positive effect on German data. Practical applications on spoken language, however, would rely on automatically predicted prosodic information. In this paper we predict pitch accents (and phrase boundaries) using a convolutional neural network (CNN) model from acoustic features extracted from the speech signal. After an assessment of the quality of these automatic prosodic annotations, we show that they also significantly improve coreference resolution.
\end{abstract}

\section{Introduction}

Noun phrase coreference resolution is the task of\blfootnote{*The two first authors contributed equally to this work.}
grouping noun phrases (NPs) together that refer to the same discourse entity in a text or
dialogue.
In Example \ref{umbach1}, taken from \newcite{Umbach}, the question for the coreference resolver, besides linking the anaphoric pronoun \textit{he} back to \textit{John}, is to decide whether \textit{an old cottage} and \textit{the shed} refer to the same entity. 

\ex. \label{umbach1}$\{$John$\}_{1}$ has $\{$an old
  cottage$\}_{2}$.\\
Last year $\{$he$\}_{1}$ reconstructed $\{$the shed$\}_{?}$.

Coreference resolution is an active NLP research area, with its own track at most NLP conferences and several shared tasks such as the CoNLL or SemEval shared tasks \cite{pradhan2012,Recasens} or the CORBON shared task 2017\footnote{\url{http://corbon.nlp.ipipan.waw.pl/}}.
Almost all work is based on text, although there exist a few systems for pronoun resolution in transcripts of spoken text \cite{StrubeMueller2003,tetrault}. It has been shown that there are differences between written and spoken text that lead to a drop in performance when coreference resolution systems developed for written text are applied on spoken text \cite{Amoia2012}.
For this reason, it may help to use additional information available from the speech signal, for example prosody.

In West-Germanic languages, such as English and German, there is a tendency for coreferent items, i.e.\ entities that have
already been introduced into the discourse (their information status is \textit{given}), to be deaccented, as the
speaker assumes the entity to be salient in the listener's discourse
model (cf.  \newcite{TerkenHirschberg1994,BaumannRiester2013,BaumannRoth2014}). We can make use of this fact by providing prosodic information to the coreference resolver.
Example~\ref{umbach2}, this time marked with prominence information, shows that prominence can help us resolve cases where the transcription is potentially ambiguous\footnote{The anaphor under consideration is typed in boldface, its antecedent
 is underlined. Accented syllables are capitalised.}.

\ex. \label{umbach2}$\{$John$\}_{1}$ has \underline{$\{$an old
  cottage$\}_{2}$}.
 \a. \label{umbach2a}Last year $\{$he$\}_{1}$ reconstructed $\{$the SHED$\}_{3}$.
\b. \label{umbach2b}Last year $\{$he$\}_{1}$ reconSTRUCted \textbf{the shed$\}_{2}$}.

The pitch accent on {\it shed} in \ref{umbach2a} leads to the interpretation that {\it the shed} and {\it the cottage} refer to different entities, where the shed is a part of the cottage (they are in a bridging relation). In contrast, in \ref{umbach2b}, {\it the shed} is deaccented, which suggests that {\it the shed} and {\it the cottage}
corefer.

A pilot study by \newcite{RoesigerRiester15} has shown that enhancing the text-based feature set for a coreference resolver, consisting of e.g. automatic part-of-speech (POS) tags and syntactic information, with pitch accents and prosodic phrasing information helps to improve coreference resolution of German spoken text. The prosodic labels used in the experiments were annotated manually, which is not only expensive but not applicable in an automatic pipeline setup.
In our paper, we present an experiment in which we replicate the main results from the pilot study by annotating the prosodic information automatically, thus omitting any manual annotations from the feature set. We show that adding prosodic information significantly helps in all of our experiments. 

\section{Prosodic features for coreference resolution}
\label{prosodicfeatures}

Similar to the pilot study, we make use of {\it pitch accents} and
{\it prosodic
phrasing}. We predict the presence of a pitch accent\footnote{We do not predict the pitch accent type (e.g. fall H*L or rise L*H) as this distinction was not helpful in the pilot study and is generally more difficult to model.}
and use phrase boundaries to derive nuclear accents, which are taken to be the last (and perceptually most prominent) accent in an intonation phrase. 
This paper tests whether previously reported tendencies for manual labels are also observable for automatic labels, namely:

\paragraph{Short NPs} Since long, complex NPs almost always have at least one pitch accent, the presence and the absence of a pitch accent is more helpful for shorter phrases.

\paragraph{Long NPs} For long, complex NPs, we  look for nuclear accents that indicate the phrase's overall prominence. If the NP contains a nuclear accent, it is assumed to be less likely to take part in coreference chains. \\

\noindent We test the following features that have proven beneficial in the pilot study. These features are derived for each NP. 

\paragraph{Pitch accent presence}
focuses on the presence of a pitch
accent, 
disregarding its type. If one accent 
 is present in the NP, this boolean feature gets assigned the
 value {\it true}, and {\it false} otherwise.

\paragraph{Nuclear accent presence} is a boolean feature comparable to
pitch accent presence. It gets assigned the value \textit{true} if
there is a nuclear accent present in the NP, \textit{false} otherwise.

\section{Data}

To ensure comparability, we use the same dataset as in the pilot study, namely the DIRNDL corpus \cite{EckartEtAl2012,BjorkelundEtAl2014}, a German radio news corpus annotated with both manual coreference and manual prosody labels. We adopt the official train, test and development split\footnote{\url{http://www.ims.uni-stuttgart.de/forschung/ressourcen/korpora/dirndl.en.html}} designed for research on coreference resolution.
The recorded news broadcasts in the DIRNDL-anaphora corpus were spoken by 13 male and 7 female speakers, in total roughly 5 hours of speech.
The prosodic annotations follow the GToBI(S) standard for pitch accent types and boundary tones and are described in \newcite{BjorkelundEtAl2014}. In this study we make use of two class labels of prosodic events: all accent types (marked by the standard ToBI *) grouped into a single class (pitch accent presence) and the same for intonational phrase boundaries (marked by \%).

\section{Automatic prosodic information}

In this section we describe the prosodic event detector used in this work. It is a binary classifier that is trained separately for either pitch accents or phrase boundaries and predicts for each word, whether it carries the respective prosodic event.

\subsection{Model}
\begin{figure}
\includegraphics[width=.45\textwidth]{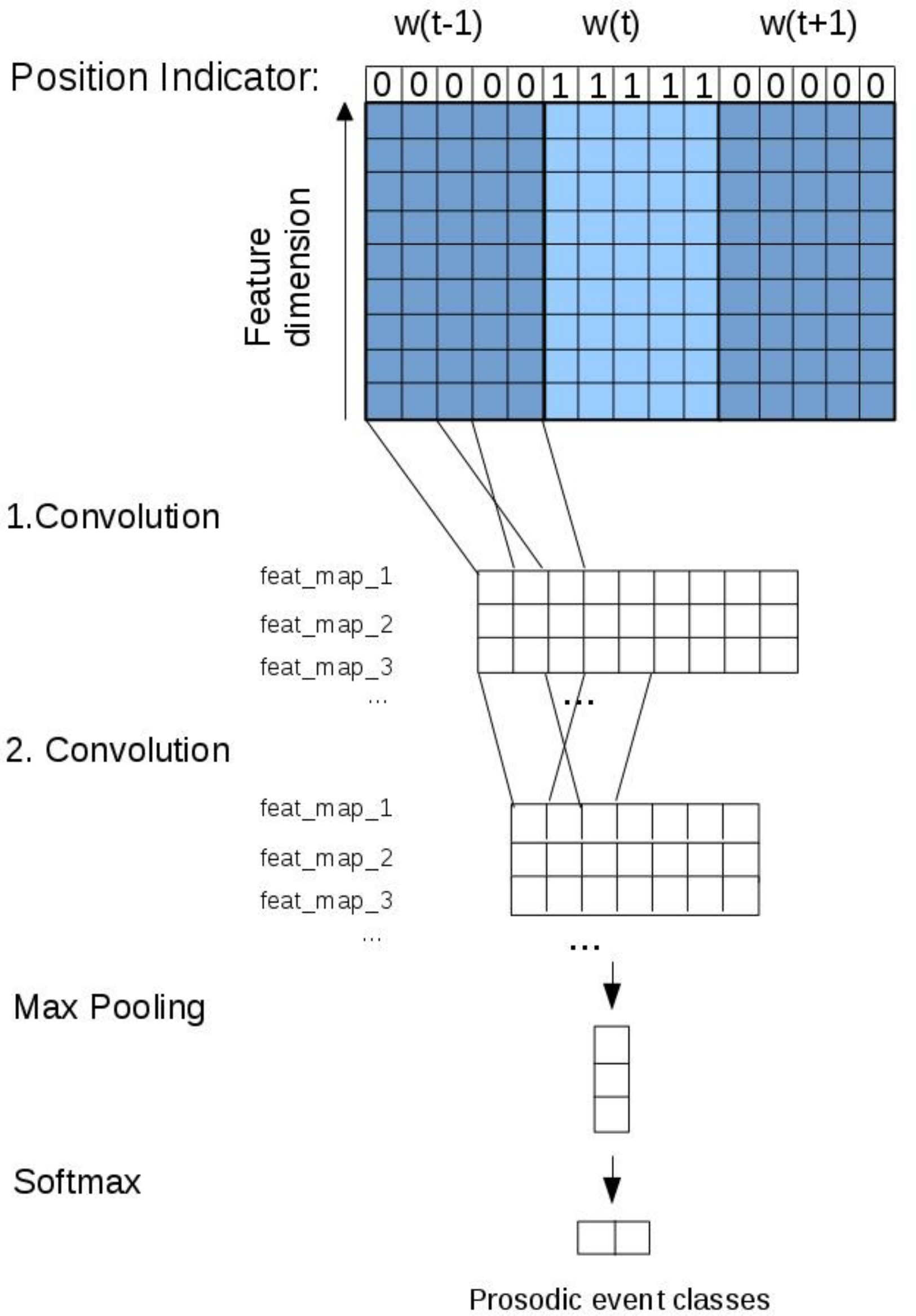}
\caption{CNN for prosodic event recognition with an input window of 3 successive words and position indicating features.}
\label{fig:model}
\end{figure}
We apply a convolutional neural network (CNN) model,  illustrated in Figure~\ref{fig:model}.
The input to the CNN is a matrix spanning the current word and its right and left context word. The input matrix is a frame-based representation of the speech signal.
The signal is divided into overlapping frames for each 20 ms with a 10 ms shift and are represented by a 6-dimensional feature vector for each frame.

We use acoustic features as well as position indicator features following \newcite{Stehwien2017} that are simple and fast to obtain.
The acoustic features were extracted from the speech signal using the OpenSMILE toolkit \cite{opensmile}.
The feature set consists of 5 features that comprise acoustic correlates of prominence: smoothed fundamental frequency (f0), RMS energy, PCM loudness, voicing probability and Harmonics-to-Noise Ratio. 
The position indicator feature is appended as an extra feature to the input matrices (see Figure~\ref{fig:model}) and aids the modelling of the acoustic context by indicating which frames belong to the current word or the neighbouring words.

We apply two convolution layers in order to expand the input information and then use max pooling to find the most salient features.
In the first convolution layer we ensure that the filters always span all feature dimensions.
All resulting feature maps are concatenated to one feature vector which is fed into the two-unit softmax layer.

\subsection{Predicting prosodic labels on DIRNDL}

We predict prosodic events for the whole DIRNDL subcorpus used in this paper. To simulate an application setting, we train the CNN model on a different dataset.
Since the acoustic correlates of prosodic events as well as the connection between sentence prosody and information status exploited in this paper are similar in English and German, we train the prosodic event detector on English data and apply the model to the German DIRNDL corpus\footnote{\newcite{Rosenberg2012} report good cross-language results of pitch accent detection on this dataset.}. The data used to train the model is a 2.5 hour subset of the Boston University Radio News Corpus \cite{Ostendorf1995} that contains speech from 3 female and 2 male speakers and that includes manually labelled pitch accents and intonational phrase boundary tones. Hence, both corpora consist of read speech by radio news anchors. The prediction accuracy on the DIRNDL anaphora corpus is 81.9\% for pitch accents and 85.5\% for intonational phrase boundary tones\footnote{The per-class accuracy is 82.1\% for pitch accents and 37.1\% for phrase boundaries. Despite these low quality phrase boundary annotations, we believe that, as a first step, their effectiveness can still be tested. This issue will be addressed in future work.}.
The speaker-independent performance of this model on the Boston dataset is 83.5\% accuracy for pitch accent detection and 89\% for phrase boundary detection. We conclude that the prosodic event detector generalises well to the DIRNDL dataset and the obtained accuracies are appropriate for our experiments.

\section{Coreference resolution}
 
In this section, we describe the coreference resolver used in our experiments and how it was applied to create the baseline system using only automatic annotations.

 \subsection{IMS HotCoref DE}
The IMS HotCoref DE coreference resolver is a state-of-the-art tool for German\footnote{\url{www.ims.uni-stuttgart.de/forschung/
ressourcen/werkzeuge/HOTCorefDe.html}} 
\cite{rosiger.2016}. It is data-driven, i.e. it learns from annotated data with the help of pre-defined features using a structured perceptron that models coreference within a document as a directed tree. This way, it can exploit the tree structure to create non-local features (features that go beyond a pair of NPs).
The standard features are text-based and consist mainly of 
string matching, part of speech, constituent parses, morphological information and combinations thereof.

\subsection{Coreference resolution using automatic preprocessing}
As we aim at coreference resolution applicable to new texts, all annotations used to create the text-based features are automatically predicted using NLP tools. 
It is frequently observed that the performance drops when the feature set is derived in this manner compared to using features based on manual annotations.
For example, the performance of IMS HotCoref DE drops from 63.61 to 48.61 CoNLL score\footnote{We report the performance of the coreference system in terms of the CoNLL score, the standard measure to assess the quality of coreference resolution.} on the reference dataset T\"uBA-9 D/Z.
The system, pre-trained on T\"uBA, yields a CoNLL score of 37.04 on DIRNDL with predicted annotations. 
This comparatively low score also confirms the assumption that the performance of a system trained on written text drops when applied to spoken text. The drop in performance can also be explained by the slightly different domains (newspaper text and radio news). However, if 
we train on the concatenation of the train and development set of DIRNDL we achieve a score of 46.11. This will serve as a baseline in the following experiments.

\section{Experiments}

We test our prosodic features by adding them to the feature set used in the baseline. We define \textit{short NPs} to be of length 3 or shorter\footnote{In our experiments, this performed even better than length 4 or shorter as used in \newcite{RoesigerRiester15}.}. In this setup, we apply the feature only to short NPs. In the \textit{all NP} setting, the feature is used for all NPs. The ratio of short vs. longer NPs in DIRNDL is roughly 3:1. Note that we evaluate on the whole test set in both cases. 
We report how the performance of the coreference resolver is affected in three settings: 

\begin{itemize}
\item[(a)]  trained and tested\\ on manual prosodic labels (short \textit{gold}), 
\item[(b)]  trained on manual prosodic labels, but tested on automatic labels (this simulates an application scenario where a pre-trained model is applied to new texts (short \textit{gold/auto}) and 
\item[(c)] trained and tested on \\automatic prosodic labels (short \textit{auto}). 

\end{itemize}

Table~\ref{pitchaccentpresence} shows the effect of the pitch accent presence feature on our data. All features perform significantly better than the baseline\footnote{We compute significance using the Wilcoxon signed rank test \cite{SiegelCastellan88} at the 0.01 level.}. As expected, the numbers are higher if we limit this feature to short NPs. We believe that this is due to the fact that the feature contributes most when it is most meaningful: on short NPs, a pitch accent makes it more likely for the NP to contain new information, whereas long NPs almost always have at least one pitch accent, regardless of its information status. We achieve the highest performance with gold labels, followed by the \textit{gold/auto} version with a score that is not significantly worse than the \textit{gold} version. This is important for applications as it suggests that the loss in performance is small when training on gold data and testing on predicted data. As expected, the version that is trained and tested on predicted data performs worse, but is still significantly better than the baseline. Hence, prosodic information is helpful in all three settings. It also shows that the assumption on short NPs in the pilot study is also true for automatic labels.

Table~\ref{nuclearaccentpresence} shows the effect of adding nuclear accent presence as a feature to the baseline. Again, we report results that are all significantly better than the baseline. The improvement is largest when we apply the feature to all NPs, i.e. also including long, complex NPs. This is in line with the findings in the pilot study for long NPs. 
If we restrict ourselves to just nuclear accents, this feature will receive the value \textit{true} for only a few of the short NPs that would otherwise have been assigned  \textit{true} in terms of general pitch accent presence.  Therefore, nuclear pitch accents do not provide sufficient information for a majority of the short NPs. For long NPs, however, the presence of a nuclear accent is more meaningful. 

 The performance of the different systems follows the pattern present for pitch accent type: \textit{gold $\textgreater$ gold/auto $\textgreater$ auto}. Again, automatic prosodic information contributes to the system's performance. 
The highest score when using automatic labels is 50.64, as compared to 53.99 with gold labels. To the best of our knowledge, these are the best results reported on the DIRNDL anaphora dataset so far.

\begin{table}[t]
\centering
\small
\begin{tabular}{l|l|l}
\hline

Baseline & 46.11\\
\hline
+ Accent &short NPs & all NPs\\
\hline
+ Presence gold& 53.99&49.68\\
+ Presence gold/auto & 52.63&50.08\\
+ Presence auto & 49.13& 49.01\\
\hline
\end{tabular}
\caption{Pitch accent presence}
\label{pitchaccentpresence}
\end{table}

\begin{table}[t]
\centering
\small
\begin{tabular}{l|l|l}
\hline

Baseline & 46.11\\
\hline

+ Nuclear accent &short NPs & all NPs\\
\hline
+ Presence gold& 48.63&52.12\\
+ Presence gold/auto & 48.46&51.45\\
+ Presence auto & 48.01&50.64\\
\hline

\end{tabular}
\caption{Nuclear accent presence}
\label{nuclearaccentpresence}
\end{table}

%

\begin{figure*}[h!]
\begin{tabular}{lllllll}
EXPERTEN & $\{$\underline{der Gro{\ss}en KOALITION}$\}_{1}$ & haben &sich &auf [...]& ein &Niedriglohn-\\
\textit{Experts} & \textit{(of) the grand coalition} & \textit{have} & \textit{themselves} & \textit{on} &\textit{a} &\textit{low wage} \\ 
\end{tabular}
\vspace*{2mm}

\begin{tabular}{lllllll}
Konzept &VERST\"ANDIGT. & Die strittigen Themen [...]& sollten &bei &der & n\"achsten \\
\textit{concept} & \textit{agreed.} & \textit{The controversial topics} & \textit{shall} & \textit{at} & \textit{the}& \textit{next}  \\
 \end{tabular}
\vspace*{2mm}

 \begin{tabular}{llll}
Spitzenrunde & \textbf{$\{$der Koalition$\}_{1}$}& ANGESPROCHEN &werden. \\
\textit{meeting} & \textit{(of) the coalition} &  \textit{raised}& \textit{be}.\\
\end{tabular}

\vspace*{3mm}
\begin{tabular}{l}
\textit{EN: Experts within the \underline{the grand coalition} have agreed on a  strategy to address [problems associated} \\ \textit{with] low income.
At the next meeting, \textbf{the coalition} will talk about the controversial issues. }
\end{tabular}
\caption{Example from the DIRNDL dataset with English translation. The candidate NP (anaphor) of the coreference chain in question is marked in boldface, the antecedent is underlined. Pitch accented words are capitalised.}
\label{example1dirndl}
\end{figure*}

\section{Analysis}
In the following section, we discuss two examples from the DIRNDL dataset that provide some insight as to how the prosodic features helped coreference resolution in our experiments. \\
The first example is shown in Figure~\ref{example1dirndl}. The coreference chain marked in this example was not predicted by the baseline version. With prosodic information, however, the fact that the NP \textit{``der Koalition''} is deaccented helped the resolver to recognise that this was given information: it refers to the recently introduced antecedent \textit{``der Gro{\ss}en Koalition''}. This effect clearly supports our assumption that the absence of pitch accents helps for short NPs.

An additional effect of adding prosodic information that we observed concerns the length of antecedents determined by the resolver. In several cases, e.g. in  Example~\ref{exampledirndl}, the baseline system incorrectly chose an embedded NP (1A) as the antecedent for a pronoun. 
The system with access to prosodic information correctly chose the longer NP (1B)\footnote{The T\"uBA-D/Z guidelines state that the maximal extension of the NP should be chosen as the markable.\\ \url{http://www.sfs.uni-tuebingen.de/fileadmin/static/ascl/resources/tuebadz-coreference-manual-2007.pdf}}. 
Our analysis confirms that this is due to
 the accent on the short NP (on \textit{Phelps}). The presence or absence of a pitch accent on the adjunct NP (on \textit{USA}) does not appear to have an impact.

\ex. \label{exampledirndl}$\{\{$Michael PHELPS$\}_{1A}$ aus den USA$\}_{1B}$. \textbf{$\{$Er$\}_{1}$} ...\\
\textit{Michael Phelps from the USA. He ...}

%
%
%
%
%

\section{Conclusion and future work}

We show that using prosodic labels that have been obtained automatically significantly improves the performance of a coreference resolver. In this work, we predict these labels using a CNN model and use these as additional features in IMS HotCoref DE, a coreference resolution system for German. Despite the quality of the predicted labels being slightly lower than the gold labels, we are still able to replicate results observed when using manually annotated prosodic information.
This encouraging result also confirms that not only is prosodic information helpful to coreference resolution, but that it also has a positive effect even when predicted by a system.

A brief analysis of the resolver's output illustrates the effect of deaccentuation.
Further work is necessary to investigate the impact on the length of the predicted antecedent.

One possibility to increase the quality of the predicted prosody labels would be to include the available lexico-syntactic information into the prosodic event detection model,
since this has been shown to improve prosodic event recognition \cite{Sun2002,Anantha2008}.
To pursue coreference resolution directly on speech, a future step would be to perform all necessary annotations on automatic speech recognition output. 
As a first step, our results on German spoken text are promising and we expect them to be generalisable to other languages with similar prosody.  

\section*{Acknowledgements}

We would like to thank Kerstin Eckart for her help with the preparation of DIRNDL data. 
This work was funded by the German Science Foundation (DFG),
Sonderforschungsbereich 732, Project A6 and A8, at the University of Stuttgart.

\bibliography{emnlp2017}

\begin{thebibliography}{}
\expandafter\ifx\csname natexlab\endcsname\relax\def\natexlab#1{#1}\fi

\bibitem[{Amoia et~al.(2012)Amoia, Kunz, and Lapshinova-Koltunski}]{Amoia2012}
Marilisa Amoia, Kerstin Kunz, and Ekaterina Lapshinova-Koltunski. 2012.
\newblock Coreference in spoken vs. written texts: a corpus-based analysis.
\newblock In {\em Proceedings of LREC\/}.

\bibitem[{Ananthakrishnan and Narayanan(2008)}]{Anantha2008}
Sankaranarayanan Ananthakrishnan and Shrikanth~S. Narayanan. 2008.
\newblock Automatic prosodic event detection using acoustic, lexical and
  syntactic evidence.
\newblock In {\em IEEE Transactions on Audio, Speech and Language
  Processing\/}. volume~16, pages 216--228.

\bibitem[{Baumann and Riester(2013)}]{BaumannRiester2013}
Stefan Baumann and Arndt Riester. 2013.
\newblock Coreference, lexical givenness and prosody in {German}.
\newblock {\em Lingua\/} 136:16--37.

\bibitem[{Baumann and Roth(2014)}]{BaumannRoth2014}
Stefan Baumann and Anna Roth. 2014.
\newblock Prominence and coreference -- {O}n the perceptual relevance of {F0}
  movement, duration and intensity.
\newblock In {\em Proceedings of Speech Prosody\/}. pages 227--231.

\bibitem[{Bj{\"o}rkelund et~al.(2014)Bj{\"o}rkelund, Eckart, Riester,
  Schauffler, and Schweitzer}]{BjorkelundEtAl2014}
Anders Bj{\"o}rkelund, Kerstin Eckart, Arndt Riester, Nadja Schauffler, and
  Katrin Schweitzer. 2014.
\newblock The extended {DIRNDL} corpus as a resource for automatic coreference
  and bridging resolution.
\newblock In {\em Proceedings of LREC\/}. pages 3222--3228.

\bibitem[{Eckart et~al.(2012)Eckart, Riester, and Schweitzer}]{EckartEtAl2012}
Kerstin Eckart, Arndt Riester, and Katrin Schweitzer. 2012.
\newblock A discourse information radio news database for linguistic analysis.
\newblock In Sebastian~Nordhoff Christian~Chiarcos and Sebastian Hellmann,
  editors, {\em Linked Data in Linguistics: Representing and Connecting
  Language Data and Language Metadata\/}, Springer, pages 65--76.

\bibitem[{Eyben et~al.(2013)Eyben, Weninger, Gro{\ss}, and
  Schuller}]{opensmile}
Florian Eyben, Felix Weninger, Florian Gro{\ss}, and Bj{\"o}rn Schuller. 2013.
\newblock Recent developments in {openSMILE}, the {M}unich open-source
  multimedia feature extractor.
\newblock In {\em Proceedings of the 21st ACM international conference on
  Multimedia\/}. pages 835--838.

\bibitem[{Ostendorf et~al.(1995)Ostendorf, Price, and
  Shattuck-Hufnagel}]{Ostendorf1995}
Mari Ostendorf, Patti Price, and Stefanie Shattuck-Hufnagel. 1995.
\newblock The {B}oston {U}niversity {R}adio {N}ews {C}orpus.
\newblock Technical Report ECS-95-001, Boston University.

\bibitem[{Pradhan et~al.(2012)Pradhan, Moschitti, Xue, Uryupina, and
  Zhang}]{pradhan2012}
Sameer Pradhan, Alessandro Moschitti, Nianwen Xue, Olga Uryupina, and Yuchen
  Zhang. 2012.
\newblock Conll-2012 shared task: Modeling multilingual unrestricted
  coreference in ontonotes.
\newblock In {\em Joint Conference on EMNLP and CoNLL-Shared Task\/}.
  Association for Computational Linguistics, pages 1--40.

\bibitem[{Recasens et~al.(2010)Recasens, M\`{a}rquez, Sapena, Mart\'{\i},
  Taul{\'e}, Hoste, Poesio, and Versley}]{Recasens}
Marta Recasens, Llu\'{\i}s M\`{a}rquez, Emili Sapena, M.~Ant\`{o}nia
  Mart\'{\i}, Mariona Taul{\'e}, V{\'e}ronique Hoste, Massimo Poesio, and
  Yannick Versley. 2010.
\newblock Semeval-2010 task 1: Coreference resolution in multiple languages.
\newblock In {\em Proceedings of the 5th International Workshop on Semantic
  Evaluation\/}. Stroudsburg, PA, USA, SemEval '10, pages 1--8.

\bibitem[{Rosenberg et~al.(2012)Rosenberg, Cooper, Levitan, and
  Hirschberg}]{Rosenberg2012}
Andrew Rosenberg, Erica Cooper, Rivka Levitan, and Julia Hirschberg. 2012.
\newblock Cross-language prominence detection.
\newblock In {\em Speech Prosody\/}.

\bibitem[{R\"osiger and Kuhn(2016)}]{rosiger.2016}
Ina R\"osiger and Jonas Kuhn. 2016.
\newblock {IMS HotCoref DE:} a data-driven co-reference resolver for {G}erman.
\newblock In {\em Proceedings of LREC 2016\/}.

\bibitem[{R{\"o}siger and Riester(2015)}]{RoesigerRiester15}
Ina R{\"o}siger and Arndt Riester. 2015.
\newblock Using prosodic annotations to improve coreference resolution of
  spoken text.
\newblock In {\em Proceedings of ACL-IJCNLP\/}. pages 83--88.

\bibitem[{Siegel and Castellan(1988)}]{SiegelCastellan88}
Sidney Siegel and N.~John~Jr. Castellan. 1988.
\newblock {\em Nonparametric Statistics for the Behavioral Sciences\/}.
\newblock McGraw-Hill, Berkeley, CA, 2nd edition.

\bibitem[{Stehwien and Vu(2017)}]{Stehwien2017}
Sabrina Stehwien and Ngoc~Thang Vu. 2017.
\newblock Prosodic event detection using convolutional neural networks with
  context information.
\newblock In {\em Proceedings of Interspeech\/}.

\bibitem[{Strube and M{\"u}ller(2003)}]{StrubeMueller2003}
Michael Strube and Christoph M{\"u}ller. 2003.
\newblock A machine learning approach to pronoun resolution in spoken dialogue.
\newblock In {\em Proceedings of the 41st Annual Meeting on Association for
  Computational Linguistics\/}. pages 168--175.

\bibitem[{Sun(2002)}]{Sun2002}
Xuejing Sun. 2002.
\newblock Pitch accent prediction using ensemble machine learning.
\newblock In {\em Proceedings of ICSLP-2002\/}. pages 16--20.

\bibitem[{Terken and Hirschberg(1994)}]{TerkenHirschberg1994}
Jacques Terken and Julia Hirschberg. 1994.
\newblock Deaccentuation of words representing `given' information: {E}ffects
  of persistence of grammatical function and surface position.
\newblock {\em Language and Speech\/} 37(2):125--145.

\bibitem[{Tetreault and Allen(2004)}]{tetrault}
Joel Tetreault and James Allen. 2004.
\newblock Dialogue structure and pronoun resolution.
\newblock In {\em Proceedings of the 5th Discourse Anaphora and Anaphor
  Resolution Colloquium\/}.

\bibitem[{Umbach(2002)}]{Umbach}
Carla Umbach. 2002.
\newblock ({D}e)accenting definite descriptions.
\newblock {\em Theoretical Linguistics\/} 2/3:251--280.

\end{thebibliography}
\bibliographystyle{emnlp_natbib}

\end{document}